%% file: main.tex
\documentclass[conference]{IEEEtran}
\usepackage{float}
\usepackage{amsmath}

\usepackage{times}
\usepackage{latexsym}
\usepackage{graphicx} 
\usepackage[T1]{fontenc}
\usepackage{lscape}
\usepackage{xcolor}
\usepackage{inconsolata}
\usepackage{booktabs}
\usepackage{graphicx}
\usepackage{placeins}
\usepackage[normalem]{ulem}
\useunder{\uline}{\ul}{}
\usepackage[hidelinks]{hyperref}
\usepackage{balance}

\ifCLASSOPTIONcompsoc
  \usepackage[nocompress]{cite}
\else
  \usepackage{cite}
\fi
\ifCLASSINFOpdf
\else
\fi
\usepackage{url}

\begin{document}
%
\title{LongKey: Keyphrase Extraction for Long Documents}

\author{\IEEEauthorblockN{Jeovane Honorio Alves\\ and Radu State}
\IEEEauthorblockA{SEDAN - SnT\\
University of Luxembourg\\
jeovane.alves,radu.state@uni.lu
}
\and
\IEEEauthorblockN{Cinthia Obladen \\de Almendra Freitas}
\IEEEauthorblockA{Graduate Program in Law (PPGD)\\ Pontifícia Universidade \\Católica do Paraná\\
cinthia.freitas@pucpr.br \\
}
\and
\IEEEauthorblockN{Jean Paul Barddal}
\IEEEauthorblockA{Graduate Program in Informatics (PPGIa)\\ Pontifícia Universidade \\Católica do Paraná\\
jean.barddal@ppgia.pucpr.br \\
}
}


%


\maketitle

\begin{abstract}
In an era of information overload, manually annotating the vast and growing corpus of documents and scholarly papers is increasingly impractical. Automated keyphrase extraction addresses this challenge by identifying representative terms within texts. However, most existing methods focus on short documents (up to 512 tokens), leaving a gap in processing long-context documents. In this paper, we introduce LongKey, a novel framework for extracting keyphrases from lengthy documents, which uses an encoder-based language model to capture extended text intricacies. LongKey uses a max-pooling embedder to enhance keyphrase candidate representation. Validated on the comprehensive LDKP datasets and six diverse, unseen datasets, LongKey consistently outperforms existing unsupervised and language model-based keyphrase extraction methods. Our findings demonstrate LongKey's versatility and superior performance, marking an advancement in keyphrase extraction for varied text lengths and domains.
\end{abstract}


%
\IEEEpeerreviewmaketitle








\input{sections/1intro}

\input{sections/2approach}

\input{sections/3experiments}
\input{sections/4results}

\input{sections/5conclusion}

\section*{Acknowledgments}
This study has been funded by the \textit{Coordenação de Aperfeiçoamento de Pessoal de Nível Superior (CAPES)} via the \textit{Programa Nacional de Cooperação Acadêmica} (PROCAD-SPFC) program.

\balance
\bibliographystyle{IEEEtran}
\bibliography{IEEEabrv, custom}
\balance
\end{document}

%% file: sections/1intro.tex
\section{Introduction}

Efficient extraction of vital information from textual documents across diverse domains is essential for effective information retrieval, especially given the vast volume of data on the internet and within organizational datasets.
In response to this need, Keyphrase Extraction (KPE) aims to identify representative keyphrases that enhance document comprehension, retrieval, and information management \cite{min2023recent, song2023survey}.

A keyword encapsulates the central theme or a distinct element of a document's subject matter. When multiple words are used, this term is referred to as a \emph{keyphrase}.
In practice, the terms \textit{keyword} and \textit{keyphrase} are often used interchangeably. This paper adopts this convention, treating \textit{keyword} and \textit{keyphrase} extraction as synonymous, applicable to terms of any length \cite{siddiqi2015keyword}.

Keyphrase extraction techniques are commonly categorized based on their underlying principles \cite{siddiqi2015keyword}. For example, unsupervised methods like TF-IDF \cite{ramos2003using} calculate term importance based on term frequency within a document and across the corpus.
RAKE \cite{rose2010automatic} assesses word relevance through co-occurrence ratios, while TextRank \cite{mihalcea2004textrank} uses a graph-based structure to measure word strength and similarity.
KeyBERT \cite{grootendorst2020keybert}, unlike unsupervised methods, employs supervised learning with pre-trained BERT embeddings \cite{devlin2018bert} and cosine similarity to determine importance and relevance. PatternRank \cite{schopf_etal_kdir22} is similar to KeyBERT, yet it uses a part-of-speech (POS) module to reduce the number of keyphrase candidates evaluated.

A recent relevant work in keyphrase extraction is JointKPE \cite{sun2021capturing}, which finetunes a BERT model for keyphrase extraction based on two strategies: global informativeness and keyphrase chunking. Different algorithms served as baselines. ChunkKPE only uses keyphrase chunking as its strategy. Likewise, RankKPE uses only global informativeness as its strategy.
TagKPE considers a five-tagging approach to facilitate n-grams extraction. And then SpanKPE, which employs a span self-attention mechanism.

HyperMatch, a new hyperbolic matching model proposed in \cite{song2022hyperbolic}, advances keyphrase extraction beyond Euclidean space, evaluating the relevance of keyphrase candidates using the Poincaré distance. The authors also combine intermediate layers of the RoBERTa \cite{liu2019roberta} model through an adaptive mixing layer to enhance representation.
Aimed in long-context documents, GELF \cite{martinez2023enhancing} is based on graph-enhanced sequence tagging, using the Longformer \cite{beltagy2020longformer} encoder. The authors constructed a text co-occurrence graph and utilized a graph convolutional network (GCN), focusing on edge prediction, to augment Longformer model embeddings. 

Although KPE is a powerful tool, most research has focused on short-context documents, such as abstracts and news articles. While many methods focus on short texts, challenges remain for longer documents. These challenges encompass diverse content structures, increased syntactic complexity, varying contexts within the same document, and limited compatibility with long-context language models. Addressing these intricacies demands developing advanced approaches explicitly tailored for the nuances of handling long-context data \cite{mahata2022ldkp, song2023survey}.

To address these challenges, in this paper, we present LongKey\footnote{Code available at \url{https://github.com/jeohalves/longkey}.}, a novel framework that extends keyphrase extraction to long documents through two key contributions. First, LongKey expands token support for encoder models like Longformer, capable of processing up to 96K tokens, ideal for inference on lengthy documents. Second, it introduces a new strategy for keyphrase candidate embedding that captures and consolidates context across the document, enabling a more accurate, context-aware extraction.

The remainder of this paper is organized as follows: Section \ref{sec:approach} details the LongKey methodology, Section \ref{sec:experiments} presents the experimental setup, Section \ref{sec:discussion} discusses the results, and Section \ref{sec:conclusion} concludes the study.

%% file: sections/2approach.tex
\section{Proposed Approach} \label{sec:approach}

\begin{figure*}[!t]
    \centering
    \includegraphics[width=.9\textwidth]{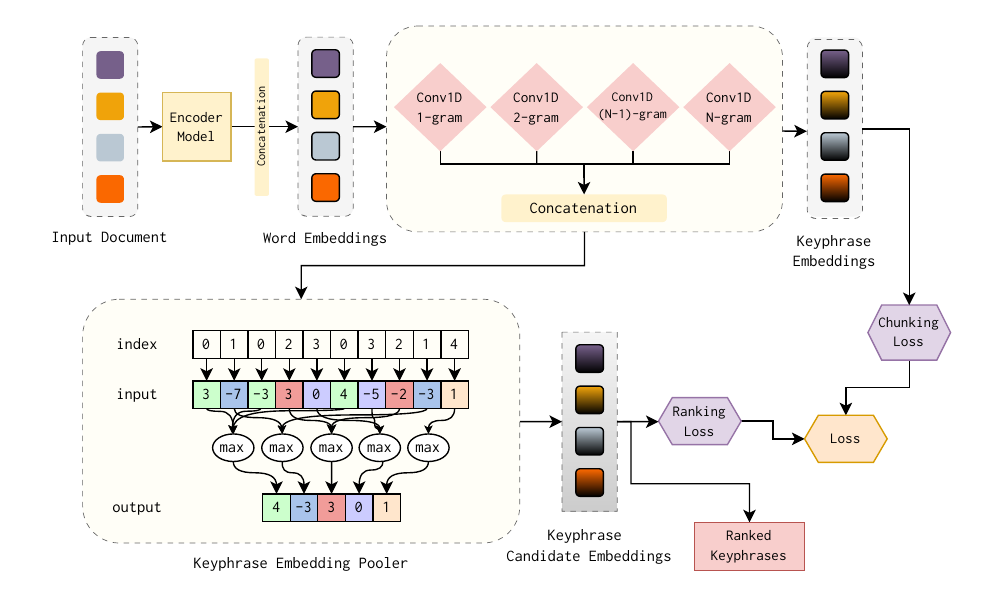}
    \caption{Overall workflow of the LongKey approach.}
    \label{fig:approach}
\end{figure*}

Our proposed methodology, dubbed \textit{LongKey}, is outlined in this section. 
LongKey operates considering three stages: initial word embedding, keyphrase candidate embedding, and candidate scoring, as shown in Figure \ref{fig:approach}. Each stage is designed to refine the selection and evaluation of keyphrases.

\subsection{Word Embedding}

To generate embeddings for long-context documents, our proposal uses the Longformer model \cite{beltagy2020longformer}. Longformer is an encoder-type language model that uniquely supports extended contexts through two innovative mechanisms: a sliding local windowed attention with a default span of 512 tokens and task-specific global attention mechanism. 

By default, each of the model's twelve attention layers produces an output embedding size of 768. Furthermore, Longformer has a positional embedding size of 4,096. We extended it to 8,192 by duplicating the same weights to the next 4,096 elements. For global attention, preliminary experiments have demonstrated optimal results by designating the initial token (\texttt{[CLS]}) as the token for global attention, i.e., the token that attends to every document token and vice-versa.

First, a tokenizer converts the input document to a numeric representation. 
Our approach uses the Longformer model as the encoder, with tokens defined by the RoBERTa \cite{liu2019roberta} tokenizer. This token representation is then processed by Longformer to generate embeddings, capturing the contextual details of each token within the document.

Even with Longformer, processing of large documents would not be possible with our current computational resources if they are not chunked. 
Therefore, documents larger than 8K tokens are split in equally sized chunks (with a maximum size of 8192 tokens). Each document is divided into chunks for processing by Longformer, and its embeddings are concatenated to create a unified representation of the entire text's tokens.

Given a document $D= \{w_1, \ldots, w_i, \ldots, w_N\}$ containing $N$ words, we use an encoder-type model to generate the token embeddings $E^T$:

\begin{equation}
    E^T = \text{Encoder}(\{w_1, \ldots, w_i, \ldots, w_N\}).
\end{equation}

The resulted operation can be represented as follows:

\begin{equation}
\begin{split}
    E^T= \{e_{1,1}, e_{1,2}, \dots, e_{1,M_1}, e_{2,1} \dots, \\ e_{i,j}, \dots, e_{N,M_N}\},
\end{split}
\end{equation}

\noindent where $e_{i,j}$ represents the embeddings of the $j$th token from the $i$th word in document $D$. Each embedding $e_{i,j}$ has a size of 768, which is omitted in the explanation for better clarity. If $N > 8192$, $D$ is grouped in chunks which are processed separately and the resulting token embeddings are concatenated together.

\subsection{Keyphrase Embedding}

Keyphrase embeddings are context-sensitive, meaning the same keyphrase can yield different embeddings based on its surrounding textual environment. Once these embeddings are crafted, they are combined into unique embeddings for each keyphrase candidate, taking into account the document's overarching thematic and semantic landscape.

Since a specific word may contain more than one token, it's necessary to create a single embedding for this word. Like JointKPE and other similar methods, we used only the first token embeddings to represent the word, since there was no significant difference between this strategy and other simple combinations evaluated, thus reducing computational calculation. These word embeddings are used as the input of our keyphrase embedding module. 

Given the token embeddings $E^T$, the word embeddings are given by preserving only the first token embeddings for each word, given as follows:

\begin{equation}
    E^W= \{e_{1,1}, e_{2,1} \dots, e_{i,1}, \dots, e_{N,1}\},
\end{equation}

\noindent which, for simplicity, we can omit the token index $j$. Then, we employ a convolutional network to construct embeddings for each potential $n$-gram keyphrase. For $n$-grams up to a predetermined maximum length, e.g., $n=5$, we use $n$ distinct 1-D convolutional layers, each with a kernel size $k$ corresponding to its $n$-gram size (i.e., $k=n$), ranging from $[1, n]$, and no padding, to generate the keyphrase embeddings from the pre-generated word embeddings. The $n$-gram representation of the keyphrase occurrence from words $w_i$ to $w_{i+k-1}$ is given by the convolutional module with kernel size $k$ is calculated as follows:

\begin{equation}
    h_{i:k}= \text{CNN}^k(\{e_{i}, \dots, e_{i+k-1}\}),
\end{equation}

\noindent where $H$, the set of keyphrase embeddings, can be represented as:

\begin{equation}
    H = \{h_{1:1}, h_{2:1}, \dots, h_{1:2}, \dots, h_{i:k}, \dots, h_{N-k, n}\}.
\end{equation}

The convolutional module generates embeddings for each keyphrase occurrence in the text. To capture the relevance of each keyphrase across the document, LongKey uses a keyphrase embedding pooler that combines all occurrences of a keyphrase candidate into a single, comprehensive representation. This approach helps emphasize the most contextually significant keyphrases.
A computationally efficient max pooling operation aggregates the diverse embeddings of the keyphrase candidate's occurrences from various text locations into a singular, comprehensive representation. Given $KP^n$ as the set the unique possible keyphrases found in $D$ with maximum size of $n$ words

\begin{equation}
    KP^n= \{kp_{1}, kp_{2}, \dots, kp_{i}, \dots, kp_{M}\},
\end{equation}  

\noindent where $M$ is the number of unique keyphrases found in $D$, from unigrams to $n$-grams, the embeddings of every occurrence of $kp_{i}$ are defined as follows:

\begin{equation}
    H^{KP_l}= \forall h_{i:k} \in H \text{ where } w_{i:k} = KP_l,
\end{equation}  

\noindent thus, for simplicity, $H^{KP_l}$ can be also represented as:

\begin{equation}
    H^{KP_l}= \{h^{l}_{1}, h^{l}_{2}, \dots, h^{l}_{i}, \dots, h^{l}_{S^{l}}\},
\end{equation} 

\noindent where $S^{l}$ is the number of occurrences in the document for a specific $KP_l$. To generate the candidate embeddings $C_l$ for the unique keyphrase $l$, a max pooling is employed as follows:

\begin{equation}
    C^{l}= \text{max}(\{h^{l}_{1}, h^{l}_{2}, \dots, h^{l}_{i}, \dots, h^{l}_{S^{l}}\}).
\end{equation} 

An overall presentation of the candidate embedding calculation is shown in the bottom-left part of Figure \ref{fig:approach}. For a clear explanation, we show an illustration with the embedding size of 1. This calculation is employed separately for each embedding position. In practice, for each keyphrase candidate, we select its occurrences and get the maximum value. In our example, the keyphrase candidate has three occurrences with values $(3,-3, 4)$ in position $j=0$. After max pooling, the value for this candidate in the position $j=0$ is $C^{i}_{j=0} = 4$. Even though this illustration only has integers, floating-point numbers are employed.

In summary, this cohesive representation encapsulates the essential details from multiple instances, facilitating a robust evaluation of keyphrase relevance for accurate ranking.  Consequently, this pooling mechanism strengthens the model’s ability to identify the most relevant keyphrases based on context, improving the precision and relevance of the extraction process.

\subsection{Candidate Scoring}

In the LongKey approach, candidate embeddings are each assigned a ranking score, with higher scores indicating keyphrases that more accurately represent the document's content. LongKey fine-tunes its performance during training by optimizing ranking and chunking losses, aligning closely with ground-truth keyphrases to ensure relevance.
For both losses, ground-truth keyphrases are positive samples. Remaining instances are considered as negative samples. 

To generate the scores for both ranking and chunking parts, we employ linear layers for different inputs. For the ranking score, we use the candidate embeddings as the input of a linear layer which converts the embedding to a single value (i.e., ranking score). Given $C^l$ as the embedding of candidate keyphrase $l$, we calculate the ranking score as follows:

\begin{equation}
    s^{l}_{\text{rank}} = \text{Linear}_{\text{rank}}(C^l)
\end{equation}

Unlike JointKPE, which might assign multiple scores to a single keyphrase candidate based on its occurrences, LongKey assigns a singular score per candidate, facilitated by the efficient proposed keyphrase embedding pooler. 

Each candidate's score is then optimized through Margin Ranking loss, enhancing the distinction between positive $y^+$ and negative $y^-$ samples by elevating the scores of the true keyphrases. This loss is defined as follows:

\begin{equation}
    \textrm{MR}_{\textrm{loss}}(s^+_{rank}, s^-_{rank}) = \max(0, -s^+_{rank} + s^-_{rank} + 1)
\end{equation}

As for the chunking score, we use the keyphrase embeddings as the input of a linear layer. Given $H^i$ as the embedding of a keyphrase $i$, we calculate the chunking score as follows:

\begin{equation}
    s^{i}_{chunk} = \text{Linear}_{\text{chunk}}(H^i)
\end{equation}

One thing to note is that LongKey maintains the same objective as JointKPE for keyphrase chunking, utilizing binary classification optimized with Cross-Entropy loss. 
Given a probability $p^+$

\begin{equation}
    p^+ = \text{Softmax}(s_{\text{chunk}})^+,
\end{equation}

\noindent representing the likelihood of a sample belonging to the positive class, and $z$, the actual binary class label of the sample (where 1 indicates positive and 0 indicates negative), the BCE loss is calculated using the formula:

\begin{equation}    
\textrm{BCE}_{\textrm{loss}} = -[z \log(p^+) + (1 - z) \log(1 - p^+)]
\end{equation}

Both losses are added together and jointly optimized across model training, similar to JointKPE. The formula is given as follows:

\begin{equation}    
\textrm{LongKey}_{\textrm{loss}} = \textrm{MR}_{\textrm{loss}} + \textrm{BCE}_{\textrm{loss}}
\end{equation}

However, distinctively, LongKey diverges from JointKPE in the objectives of its loss functions. Regarding the ranking loss function, LongKey is specifically designed to refine the embeddings of keyphrase candidates, in contrast to JointKPE's focus on optimizing the embeddings of individual keyphrase instances, thereby enhancing the model's overall precision and contextual sensitivity in keyphrase extraction.

%% file: sections/3experiments.tex
\section{Experimental Setup} \label{sec:experiments}

This section outlines the empirical evaluation of the LongKey method, providing a comprehensive overview of the experimental datasets and the specific configurations underpinning our analysis. 

\subsection{Datasets}

Robust and large datasets must be employed to train language models and evaluate the capability of an approach in extracting relevant keyphrases from an input document. 
Many large datasets typically only contain the title and abstract of scientific papers. They are sub-optimal in evaluating long context-based keyphrase extractors since they generally have samples with less than 512 tokens.

Due to the scarcity of datasets containing a high volume of lengthy documents, the \emph{Long Document Keyphrase Identification Dataset} (LDKP) was formulated specifically for extracting keyphrases from full-text papers (which generally surpass 512 tokens) \cite{mahata2022ldkp}. LDKP has two datasets: 

\textbf{LDKP3K}: A variation of the KP20K dataset \cite{meng2017deep}, which contains approximately 100 thousand samples and an average of 6027 words per document.

\textbf{LDKP10K}: A variation of the OAGKx dataset \cite{ccano2020two}, containing more than 1.3M documents, averaging 4384 words per sample.

Other datasets are employed in a zero-shot fashion, i.e., inference only, to assess the capability of different methods trained on both datasets to adapt to different domains and patterns. These datasets are the following:

\textbf{Krapivin} \cite{krapivin2009large}: Features 2,304 full scientific papers from the computer science domain published by ACM.

\textbf{SemEval2010} \cite{kim2010semeval}: Comprises 244 ACM scientific papers across four distinct sub-domains: distributed systems; information search and retrieval; distributed artificial intelligence -- multiagent systems; social and behavioral sciences -- economics.

\textbf{NUS} \cite{nguyen2007keyphrase}: This dataset contains 211 scientific conference papers with keyphrases annotated by student volunteers, offering a unique perspective on keyphrase relevance.

\textbf{FAO780} \cite{medelyan2008domain}: With 780 documents from the agricultural sector labeled by FAO staff using the AGROVOC thesaurus, this dataset tests the models' performance on domain-specific terminology.

\textbf{NLM500}  \cite{aronson2000nlm}: This collection of 500 biomedical papers, annotated with terms from the MeSH thesaurus, assesses the methods' capability in the biomedical domain.

\textbf{TMC} \cite{kontostathis2010text}: Including 281 chat logs related to child grooming from the Perverted Justice project, this dataset, with documents and keyphrases based on the formatting from \cite{alves2023detecting}, introduces the challenge of informal text and sensitive content.

Although the focus is on long-context documents, it's possible to use the evaluated methods on short documents. To assess the effectiveness of the models trained on the LDKP datasets, we evaluate them on two of the most popular short-context datasets: KP20k and OpenKP:

\textbf{KP20k} \cite{meng2017deep}: Highly correlated with the LDKP3K dataset, the KP20k is a dataset containing more than 500 thousand abstracts of scientific papers (20 thousand of abstracts for the validation and test subsets each).

\textbf{OpenKP} \cite{xiong2019open}: The OpenKeyPhrase (OpenKP) is a popular short-context dataset containing more than 140 thousand of real-world web documents, where their keyphrases were human-annotated.  

\subsection{Experimental Settings}

Our experiments utilized two NVIDIA RTX 3090 GPUs, with 24GB VRAM each. The training regimen was guided by the AdamW optimizer, combined with a cosine annealing learning rate scheduler, with a learning rate value of $5\times10^{-5}$, and warm-up for the initial 10\% training iterations. To circumvent VRAM constraints, we employed gradient accumulation, achieving an effective batch size of 16 in the training phase. 
To maintain clarity and consistency in our reporting, we use the terms ``iterations'' and ``gradient updates'' interchangeably. 

We set a maximum token limit of 8,192 during training to accommodate the length of the documents within our available computational resources. The positional embedding was expanded to 8,192, duplicating the original size used by Longformer, which enhances support for longer chunks in inference mode (tested up to 96K in total). We limit keyphrases to a maximum of five words ($k=[1,5]$) to maintain computational efficiency and align with standard practices in keyphrase extraction. In the evaluation, longer ground-truth keyphrases are considered as false negatives.
Moreover, models were trained on LDKP3K for 25 thousand iterations. Since LDKP10K had a substantially higher number of samples, we trained it for 78,125 iterations (i.e., almost an entire epoch).
We also evaluated some methods with the BERT model, where we also used chunking to extend training to 8,192 tokens.

To maintain consistency in our analysis and ensure fair comparisons, we used the Longformer model for all supervised approaches that are encoder-based and fine-tuned on the LDKP datasets. Moreover, we employed the same global attention mask as used in LongKey. 

Model performance was quantitatively assessed using the F1-score, the harmonic mean between precision and recall, for the most significant K keyphrase candidates (F1@K), with K's value determined based on the overall average of keyphrases per document in each dataset, also following choices of related works, e.g., \cite{kong2023promptrank}. 
Given 

\begin{equation}
    \hat Y = [\hat y_1, \hat y_2, \dots, \hat y_M]
\end{equation}

\noindent as the predicted keyphrases sorted by their ranking scores in a decreasing order, and $Y$ as the ground-truth keyphrases of a given document (with no specific order), we can calculate the F1-score and its intermediary metrics, i.e., precision and recall; using the top-K predicted keyphrases, given by 

\begin{equation}
    \hat Y_{:k} = [\hat y_1, \hat y_2, \dots, \hat y_{min(K,M)}].
\end{equation}

We calculate the intermediary metrics as follows:

\begin{equation}
    \text{Precision}@K = \frac{|\hat Y_{:k} \cap Y|}{|\hat Y_{:k}|}, \text{Recall}@K = \frac{|\hat Y_{:k} \cap Y|}{|Y|}, 
\end{equation}

\noindent then, with these two metrics, we calculate the F1-score at the top-K keyphrases as follows:

\begin{equation}
    \text{F1}@K = 2 \times \frac{\text{Precision}@K \times \text{Recall}@K}{\text{Precision}@K + \text{Recall}@K}.
\end{equation}

Another relevant metric, proposed in \cite{yuan2018one}, is a variation of the F1@K defined as F1@$\mathcal{O}$. Here, $\mathcal{O}$ is the number of ground-truth keyphrases (i.e., oracle), thus $K = |Y|$, which is dynamically calculated depending on the document. This metric is independent to the method's output, given the effectiveness of each method only with the needed predicted keyphrases. 

We also employed an additional evaluation: F1@Best. Basically, we evaluate which is the $K$ that have the best harmonic mean between recall and precision, i.e., best F1-score. 
The purpose of this additional evaluation is to verify how far is the optimum $K$ for a specific method in a specific dataset is from the selected $K$s. We put a threshold of $K \leq 100$ to not deviate strongly from the default $K$s. 

Furthermore, we employed the Porter Stemmer, from the NLTK package \cite{loper2002nltk}, for all experiments, but no lemmatization was applied. 
Stemming was applied for both candidate and ground-truth keyphrases. Duplicated ground-truth keyphrases were cleaned, removing the possibility of duplicated keyphrases erroneously improving the F1-score.

%% file: sections/4results.tex
\section{Results and Discussion} \label{sec:discussion}

In this section, we delve into the performance outcomes on two primary datasets, extending our analysis to encompass zero-shot learning scenarios and domain-shift adaptability. Moreover, we unravel the contribution of the keyphrase embedding pooler, performance estimation, and inference on short-context documents.

\subsection{LDKP Datasets}

Table \ref{tab:ldkp_results} presents the comparative results on the LDKP3K test subset, encompassing both unsupervised methods and models finetuned on the LDKP3K and LDKP10K training subsets. It's noteworthy that, aside from GELF, a standard benchmark model, all fine-tuned methods are tailored adaptations designed to handle extensive texts, utilizing the BERT (only when trained on LDKP3K) and Longformer architecture for enhanced context processing. Our approach was also evaluated without chunking, i.e., max of 8192 tokens, identified as \textit{LongKey8K}. 

\begin{table*}[h!tb]
\centering
\caption{Results obtained on LDKP test subsets. Values in \%. The best scores for each K are in \textbf{bold}. Best scores only in a specific section are \underline{underlined}. * GELF score was reported in its paper without a specific K value.}
\label{tab:ldkp_results}
\begin{tabular}{@{}lrrrrrrrrrr@{}}
\toprule
 & \multicolumn{5}{c}{\textbf{LDKP3K}} & \multicolumn{5}{c}{\textbf{LDKP10K}} \\ \midrule
\textbf{F1@K} & \textbf{@4} & \textbf{@5} & \textbf{@6} & \textbf{$\mathcal{O}$} & \textbf{@Best} & \textbf{@4} & \textbf{@5} & \textbf{@6} & \textbf{$\mathcal{O}$} & \textbf{@Best} \\ \midrule
TF-IDF & {\ul 8.64} & {\ul 9.08} & {\ul 9.40} & {\ul 8.75} & {\ul 9.72@9} & {\ul 7.45} & {\ul 7.88} & {\ul 8.12} & {\ul 7.77} & {\ul 8.41@9} \\
TextRank & 6.28 & 6.90 & 7.19 & 6.68 & 8.01@12 & 5.11 & 5.47 & 5.82 & 5.48 & 6.54@14 \\
PatternRank & 7.50 & 8.24 & 8.56 & 7.33 & 8.65@8 & 5.62 & 6.13 & 6.46 & 6.12 & 7.23@14 \\ \midrule
\multicolumn{11}{c}{\textbf{Trained on LDKP3K}} \\ \midrule
GELF* & - & - & - & 27.10 & - & - & - & - & - & - \\
SpanKPE & 30.27 & 30.08 & 29.43 & 31.08 & 30.27@4 & 19.99 & 20.37 & 20.39 & 21.00 & 20.39@6 \\
TagKPE & 34.50 & 34.52 & 33.94 & 36.58 & 34.52@5 & 21.48 & 21.84 & 21.92 & 22.56 & 21.92@6 \\
ChunkKPE & 31.43 & 31.17 & 30.55 & 32.81 & 31.43@4 & 20.12 & 20.45 & 20.50 & 21.06 & 20.50@6 \\
RankKPE & 36.83 & 36.61 & 35.81 & 38.38 & 36.83@4 & 23.14 & 23.70 & 23.84 & 24.31 & 23.84@6 \\
JointKPE & {\ul 37.50} & {\ul 37.23} & {\ul 36.54} & {\ul 39.41} & {\ul 37.50@4} & {\ul 23.67} & {\ul 24.23} & {\ul 24.37} & {\ul 24.98} & {\ul 24.37@6} \\
HyperMatch & 36.34 & 36.37 & 35.78 & 38.23 & 36.37@5 & 23.20 & 23.64 & 23.77 & 24.20 & 23.77@6 \\ \midrule
BERT-SpanKPE & 29.80 & 30.00 & 29.51 & 31.08 & 30.00@5 & 20.94 & 21.46 & 21.50 & 21.97 & 21.50@6 \\
BERT-TagKPE & 34.13 & 34.15 & 33.49 & 36.09 & 34.15@5 & 21.03 & 21.40 & 21.40 & 21.87 & 21.40@5 \\
BERT-ChunkKPE & 31.80 & 31.77 & 31.35 & 33.89 & 31.80@4 & 19.19 & 19.68 & 19.74 & 20.36 & 19.74@6 \\
BERT-RankKPE & 36.28 & 36.43 & 35.53 & 38.38 & 36.43@5 & 23.32 & 23.77 & 23.89 & 24.35 & 23.89@6 \\
BERT-JointKPE & {\ul 37.19} & {\ul 37.28} & {\ul 36.59} & {\ul 39.94} & {\ul 37.28@5} & {\ul 23.66} & {\ul 24.25} & {\ul 24.26} & {\ul 25.08} & {\ul 24.26@6} \\
BERT-HyperMatch & 36.17 & 36.31 & 35.49 & 38.27 & 36.31@5 & 23.63 & 24.10 & 24.16 & 24.74 & 24.16@6 \\ \midrule
\textbf{LongKey} & 39.50 & 39.50 & \textbf{38.57} & \textbf{41.84} & 39.50@5 & 25.17 & 25.78 & 25.77 & 26.45 & 25.78@5 \\
\textbf{BERT-LongKey} & 38.67 & 38.68 & 37.98 & 40.43 & 38.68@5 & {\ul 25.36} & {\ul 26.00} & {\ul 26.10} & {\ul 26.58} & {\ul 26.10@6} \\
\textbf{LongKey8K} & \textbf{39.55} & \textbf{39.54} & \textbf{38.57} & \textbf{41.84} & \textbf{39.55@4} & 25.15 & 25.75 & 25.77 & 26.50 & 25.77@6 \\ \midrule
\multicolumn{11}{c}{\textbf{Trained on LDKP10K}} \\ \midrule
SpanKPE & 25.83 & 25.81 & 25.49 & 26.54 & 25.83@4 & 32.17 & 32.21 & 31.75 & 34.90 & 32.21@5 \\
TagKPE & 30.06 & 30.12 & 29.58 & 31.48 & 30.12@5 & 41.12 & 40.68 & 39.64 & 46.47 & 41.12@4 \\
ChunkKPE & 23.93 & 23.70 & 23.11 & 24.65 & 23.93@4 & 36.22 & 35.42 & 34.43 & 40.55 & 36.22@4 \\
RankKPE & 28.20 & 28.39 & 28.08 & 29.04 & 28.39@5 & 37.98 & 38.23 & 37.89 & 42.37 & 38.23@5 \\
JointKPE & 29.79 & 29.78 & 29.44 & 30.61 & 29.79@4 & 39.86 & 39.95 & 39.45 & 44.73 & 39.95@5 \\
HyperMatch & 27.98 & 28.21 & 28.07 & 29.11 & 28.21@5 & 37.44 & 37.52 & 37.25 & 41.67 & 37.52@5 \\ 
\textbf{LongKey} & {\ul 31.84} & {\ul 31.94} & {\ul 31.69} & {\ul 32.57} & {\ul 31.94@5} & \textbf{41.57} & \textbf{41.81} & \textbf{41.00} & \textbf{47.26} & \textbf{41.81@5} \\ \bottomrule
\end{tabular}
\end{table*}

Among the evaluated methods, LongKey8K emerged as the best, achieving an F1@5 of 39.55\% and F1@$\mathcal{O}$ of 41.84\%. Remarkably, even under a domain shift when trained on the broader LDKP10K dataset, which includes a more comprehensive array of topics beyond computer science, LongKey maintained its lead with an F1@5 of 31.94\% and F1@$\mathcal{O}$ of 32.57\%.

Performance metrics on the LDKP10K test subset are also provided in Table \ref{tab:ldkp_results}, where LongKey emerges as the leading method, achieving an F1@5 of 41.81\%. 

While LongKey trained on the LDKP3K dataset outperformed other models trained on the same dataset, it scored significantly lower when compared to its performance on the LDKP10K dataset, indicative of dataset-specific variations in effectiveness. This discrepancy, especially the reduced efficacy on the LDKP10K subset, could be attributed to the significant skew towards computer science papers within the LDKP3K dataset, as detailed in the LDKP study.

Generally, the evaluated methods had superior F1@$\mathcal{O}$ than F1s at specific Ks, suggesting that, for the LDKP datasets, ground-truth keyphrases were ranked higher in prediction.

\subsection{Unseen Datasets}

Without any finetuning on their respective data, LongKey and related methods were evaluated across six diverse domains, as shown in Tables \ref{tab:unseen_datasets1} and \ref{tab:unseen_datasets2}.
Remarkably, LongKey outperformed other methods in nearly all tested datasets, with the exception of SemEval2010 and TMC where its results were slightly below the top performers (HyperMatch and RankKPE, respectively).

\begin{table*}[h!tb]
\centering
\caption{Results obtained in unseen datasets with models trained on LDKP3K and LDKP10K training subsets. Values in \%. Best scores, for each K and dataset, are in \textbf{bold}. Best scores only in a specific section are \underline{underlined}. * GELF scores were reported in its paper without a specific K value.}
\label{tab:unseen_datasets1}
\begin{tabular}{@{}lrrrrrrrrrrrrrrr@{}}
\toprule
\multicolumn{13}{c}{\textbf{Unseen datasets}} \\ \midrule
 & \multicolumn{4}{c}{\textbf{Krapivin}} & \multicolumn{4}{c}{\textbf{SemEval2010}} & \multicolumn{4}{c}{\textbf{NUS}} \\ \midrule
\textbf{F1@K} & \textbf{@4} & \textbf{@5} & \textbf{@6} & \textbf{@$\mathcal{O}$} & \textbf{@5} & \textbf{@10} & \textbf{@15} & \textbf{@$\mathcal{O}$} & \textbf{@5} & \textbf{@10} & \textbf{@15} & \textbf{@$\mathcal{O}$} \\ \midrule
TF-IDF & 6.30 & 7.02 & 7.45 & 6.40 & {\ul 6.62} & 8.80 & 10.07 & 9.42 & {\ul 10.44} & {\ul 12.22} & {\ul 12.38} & {\ul 11.98} \\
TextRank & 4.87 & 5.26 & 5.77 & 5.23 & 6.53 & {\ul 8.95} & {\ul 10.11} & {\ul 9.54} & 7.83 & 10.63 & 11.73 & 9.47 \\
PatternRank & {\ul 6.72} & {\ul 7.17} & {\ul 7.61} & {\ul 6.81} & 6.24 & 7.91 & 9.08 & 8.16 & 8.53 & 9.89 & 11.15 & 10.22 \\ \midrule
\multicolumn{13}{c}{\textbf{Trained on LDKP3K}} \\ \midrule
GELF* & - & - & - & - & - & 16.70 & - & - & - & 21.50 & - & - \\
SpanKPE & 27.59 & 27.62 & 27.22 & 28.62 & 20.78 & 24.81 & 25.42 & 25.72 & 29.68 & 30.47 & 28.30 & 33.04 \\
TagKPE & 29.87 & 29.72 & 29.32 & 31.01 & 21.81 & 24.72 & 25.14 & 25.57 & 28.78 & 31.25 & 29.09 & 32.12 \\
ChunkKPE & 27.90 & 27.74 & 27.50 & 28.89 & 20.32 & 23.58 & 23.73 & 24.29 & 27.77 & 28.66 & 26.84 & 30.46 \\
RankKPE & 32.00 & 31.82 & 31.19 & 33.32 & 20.43 & 24.99 & 25.22 & 25.53 & 29.22 & 31.64 & 30.30 & 33.32 \\
JointKPE & {\ul 32.55} & {\ul 32.42} & {\ul 32.10} & {\ul 33.73} & 19.08 & 25.10 & 25.73 & 25.80 & 28.22 & 31.12 & 30.54 & 33.61 \\
HyperMatch & 31.22 & 31.44 & 31.27 & 32.79 & {\ul 22.20} & {\ul 26.64} & {\ul 26.75} & {\ul 26.82} & {\ul 31.27} & {\ul 33.53} & {\ul 32.23} & {\ul 35.14} \\ \midrule
BERT-SpanKPE & 27.18 & 27.16 & 26.82 & 28.15 & 20.78 & 25.50 & 25.63 & 26.45 & {\ul 29.91} & 30.96 & 28.34 & 31.30 \\
BERT-TagKPE & 26.20 & 26.30 & 25.85 & 27.33 & 19.00 & 22.41 & 22.63 & 22.53 & 27.51 & 27.81 & 26.46 & 30.43 \\
BERT-ChunkKPE & 24.79 & 24.67 & 24.38 & 25.72 & 18.35 & 21.93 & 22.13 & 22.61 & 26.32 & 27.70 & 26.71 & 27.70 \\
BERT-RankKPE & 31.20 & 31.43 & 31.04 & 32.49 & 20.38 & 24.95 & 25.94 & 25.94 & 26.07 & 30.05 & 29.59 & 30.95 \\
BERT-JointKPE & 32.06 & {\ul 32.17} & {\ul 31.80} & 33.45 & 22.45 & 26.09 & 25.68 & 26.91 & 26.57 & 30.34 & 29.62 & 31.06 \\
BERT-HyperMatch & {\ul 32.16} & 32.14 & 31.79 & {\ul 33.47} & \textbf{24.35} & \textbf{27.62} & {\ul 26.85} & \textbf{27.85} & 28.98 & {\ul 31.82} & {\ul 31.08} & {\ul 33.27} \\ \midrule
\textbf{LongKey} & \textbf{34.96} & \textbf{34.82} & 34.21 & \textbf{36.31} & {\ul 22.31} & {\ul 26.36} & \textbf{27.37} & {\ul 27.74} & 30.02 & \textbf{33.32} & \textbf{32.51} & \textbf{34.95} \\
\textbf{BERT-LongKey} & 34.67 & 34.86 & 34.30 & 36.07 & 19.93 & 24.06 & 25.34 & 25.69 & 24.46 & 28.60 & 29.34 & 29.43 \\
\textbf{LongKey8K} & 34.94 & 34.85 & \textbf{34.23} & 36.29 & {\ul 22.31} & {\ul 26.36} & 27.31 & 27.60 & \textbf{30.09} & 33.19 & 32.47 & \textbf{34.95} \\ \midrule
\multicolumn{13}{c}{\textbf{Trained on LDKP10K}} \\ \midrule
SpanKPE & 24.63 & 25.13 & 24.91 & 25.52 & 22.02 & 25.35 & 26.17 & 26.29 & 26.00 & 28.19 & 26.48 & 29.56 \\
TagKPE & 26.22 & 26.57 & 26.38 & 27.43 & 21.54 & {\ul 25.82} & 26.02 & 26.59 & 25.86 & 27.16 & 26.63 & 29.13 \\
ChunkKPE & 21.37 & 21.50 & 21.23 & 22.30 & 18.57 & 20.97 & 20.54 & 20.80 & 24.56 & 26.11 & 24.08 & 26.85 \\
RankKPE & 25.56 & 26.05 & 26.15 & 26.88 & 16.47 & 20.58 & 22.59 & 22.06 & 25.18 & 26.57 & 26.34 & 27.78 \\
JointKPE & 26.68 & 27.04 & 27.11 & 27.68 & 18.23 & 21.69 & 23.23 & 23.02 & 25.43 & 26.42 & 25.76 & 27.81 \\
HyperMatch & 25.23 & 25.70 & 26.01 & 26.65 & 16.94 & 21.11 & 23.37 & 23.26 & 24.50 & 26.08 & 25.60 & 27.16 \\
\textbf{LongKey} & {\ul 29.90} & {\ul 30.52} & {\ul 30.20} & {\ul 31.33} & {\ul 22.26} & 25.77 & {\ul 26.61} & {\ul 26.79} & {\ul 27.93} & {\ul 29.20} & {\ul 28.06} & {\ul 30.34} \\ \bottomrule
\end{tabular}
\end{table*}

\begin{table*}[h!tb]
\centering
\caption{Results obtained in unseen datasets with models trained on LDKP3K and LDKP10K training subsets. Values in \%. Best scores, for each K and dataset, are in \textbf{bold}. Best scores only in a specific section are \underline{underlined}.}
\label{tab:unseen_datasets2}
\begin{tabular}{@{}lrrrrrrrrrrrrrrr@{}}
\toprule
\multicolumn{13}{c}{\textbf{Unseen datasets}} \\ \midrule
 & \multicolumn{4}{c}{\textbf{FAO780}} & \multicolumn{4}{c}{\textbf{NLM500}} & \multicolumn{4}{c}{\textbf{TMC}} \\ \midrule
\textbf{F1@K} & \textbf{@4} & \textbf{@5} & \textbf{@6} & \textbf{@$\mathcal{O}$} & \textbf{@5} & \textbf{@10} & \textbf{@15} & \textbf{@$\mathcal{O}$} & \textbf{@40} & \textbf{@50} & \textbf{@60} & \textbf{@$\mathcal{O}$} \\ \midrule
TF-IDF & 7.21 & 7.97 & 8.31 & 8.37 & {\ul 4.66} & {\ul 5.69} & {\ul 5.90} & {\ul 5.61} & 1.92 & 2.14 & 2.41 & 2.09 \\
TextRank & {\ul 9.62} & {\ul 10.23} & {\ul 10.65} & {\ul 10.95} & 4.30 & 5.35 & 5.82 & 5.32 & 4.83 & 5.70 & 6.30 & 5.35 \\
PatternRank & 1.39 & 1.62 & 1.84 & 1.86 & 2.00 & 3.21 & 3.55 & 2.71 & {\ul 6.91} & {\ul 7.27} & {\ul 7.45} & {\ul 6.93} \\ \midrule
\multicolumn{13}{c}{\textbf{Trained on LDKP3K}} \\ \midrule
SpanKPE & 15.42 & 16.23 & 16.34 & 16.83 & 11.15 & 12.29 & 11.94 & 12.34 & 12.10 & 13.04 & 13.63 & 12.58 \\
TagKPE & 18.85 & 19.31 & 19.29 & 20.42 & 13.57 & 13.85 & 13.01 & 14.34 & 14.85 & 15.63 & 16.24 & 15.15 \\
ChunkKPE & 16.53 & 17.13 & 17.41 & 17.89 & 11.85 & 12.52 & 12.15 & 12.77 & 13.63 & 14.37 & 14.99 & 13.94 \\
RankKPE & 19.34 & 19.87 & 20.42 & 20.64 & 14.08 & 14.78 & 14.11 & 15.12 & {\ul 16.21} & {\ul 17.09} & {\ul 17.62} & {\ul 16.53} \\
JointKPE & 19.35 & {\ul 19.88} & 19.98 & 20.19 & {\ul 14.16} & {\ul 15.14} & {\ul 14.52} & {\ul 15.37} & 15.26 & 16.11 & 16.65 & 15.38 \\
HyperMatch & {\ul 19.50} & 19.81 & {\ul 20.23} & {\ul 20.76} & 13.64 & 14.38 & 13.91 & 14.59 & 15.50 & 16.02 & 16.16 & 15.89 \\ \midrule
BERT-SpanKPE & 16.08 & 16.45 & 17.03 & 17.26 & 11.97 & 12.35 & 12.04 & 12.81 & 15.26 & 16.28 & 16.64 & 15.92 \\
BERT-TagKPE & 17.22 & 17.77 & 17.82 & 18.10 & 12.88 & 13.56 & 13.38 & 14.34 & 13.50 & 14.48 & 15.10 & 13.86 \\
BERT-ChunkKPE & 13.96 & 14.49 & 14.59 & 14.10 & 11.90 & 12.32 & 11.82 & 12.42 & 13.78 & 14.57 & 14.94 & 14.43 \\
BERT-RankKPE & 17.26 & 18.68 & 19.42 & 19.25 & 13.43 & 13.98 & 13.75 & 14.13 & \textbf{16.80} & \textbf{17.44} & \textbf{17.78} & \textbf{17.75} \\
BERT-JointKPE & 17.58 & 18.74 & 18.99 & 19.29 & {\ul 14.74} & {\ul 14.64} & {\ul 14.11} & {\ul 15.27} & 15.86 & 16.51 & 17.05 & 16.71 \\
BERT-HyperMatch & {\ul 18.77} & {\ul 19.25} & {\ul 19.35} & {\ul 20.16} & 13.11 & 14.32 & 13.70 & 14.72 & 15.23 & 16.31 & 16.81 & 16.09 \\ \midrule
\textbf{LongKey} & 20.90 & 21.70 & 21.87 & 22.34 & 14.24 & 14.96 & 14.21 & 15.41 & 15.89 & 16.43 & 16.75 & 16.20 \\
\textbf{BERT-LongKey} & \textbf{22.20} & \textbf{22.93} & \textbf{22.67} & \textbf{23.18} & {\ul 14.94} & {\ul 15.80} & {\ul 15.04} & {\ul 16.12} & {\ul 16.69} & {\ul 17.31} & {\ul 17.68} & {\ul 17.13} \\
\textbf{LongKey8K} & 20.91 & 21.77 & 21.84 & 22.23 & 14.25 & 15.00 & 14.21 & 15.35 & 15.92 & 16.43 & 16.75 & 16.26 \\ \midrule
\multicolumn{13}{c}{\textbf{Trained on LDKP10K}} \\ \midrule
SpanKPE & 17.40 & 18.07 & 18.12 & 18.45 & 15.00 & 16.68 & 16.49 & 16.63 & 11.50 & 12.04 & 12.31 & 11.82 \\
TagKPE & 19.89 & 20.72 & 20.69 & 21.47 & 16.23 & 17.57 & 17.04 & 17.52 & {\ul 12.09} & {\ul 12.98} & {\ul 13.75} & {\ul 12.31} \\
ChunkKPE & 13.17 & 13.18 & 13.00 & 14.17 & 13.27 & 14.13 & 13.01 & 14.56 & 1.20 & 1.53 & 1.83 & 0.82 \\
RankKPE & 18.11 & 19.01 & 19.45 & 19.77 & 15.96 & 18.94 & 18.64 & \textbf{18.86} & 8.53 & 9.47 & 10.13 & 9.01 \\
JointKPE & 18.03 & 19.05 & 19.54 & 19.88 & 16.24 & 17.92 & 17.67 & 17.96 & 9.47 & 10.49 & 10.80 & 9.67 \\
HyperMatch & 17.98 & 18.74 & 18.95 & 19.63 & 14.96 & 18.43 & 18.53 & 18.00 & 9.95 & 10.97 & 11.65 & 10.33 \\
\textbf{LongKey} & {\ul 20.00} & {\ul 21.02} & {\ul 21.20} & {\ul 21.92} & \textbf{16.49} & \textbf{19.19} & \textbf{18.78} & \textbf{18.86} & 10.87 & 11.51 & 11.80 & 10.88 \\ \bottomrule
\end{tabular}
\end{table*}

The choice of LDKP training dataset--LDKP3K or LDKP10K--significantly influenced performance across the unseen datasets, with LDKP3K-trained models excelling in every dataset with the exception of the NLM500 dataset. Although LDKP10K had broader areas of study, LDPK3K had overall longer samples, with an average of 6,027 words per document against an average of 4,384 words in the LDKP10K. 
Further studies are encouraged to assess the influence of study areas and sample size. 

Another thing to note is that, for the unseen datasets, there was a balance dispute between BERT and Longformer-based methods as the best one, even for LongKey. Although access the robustness of BERT with a chunking approach, it also show room for improvements regarding long-context encoders.

\begin{figure*}[h!tb]
    \centering
    \includegraphics[width=\linewidth]{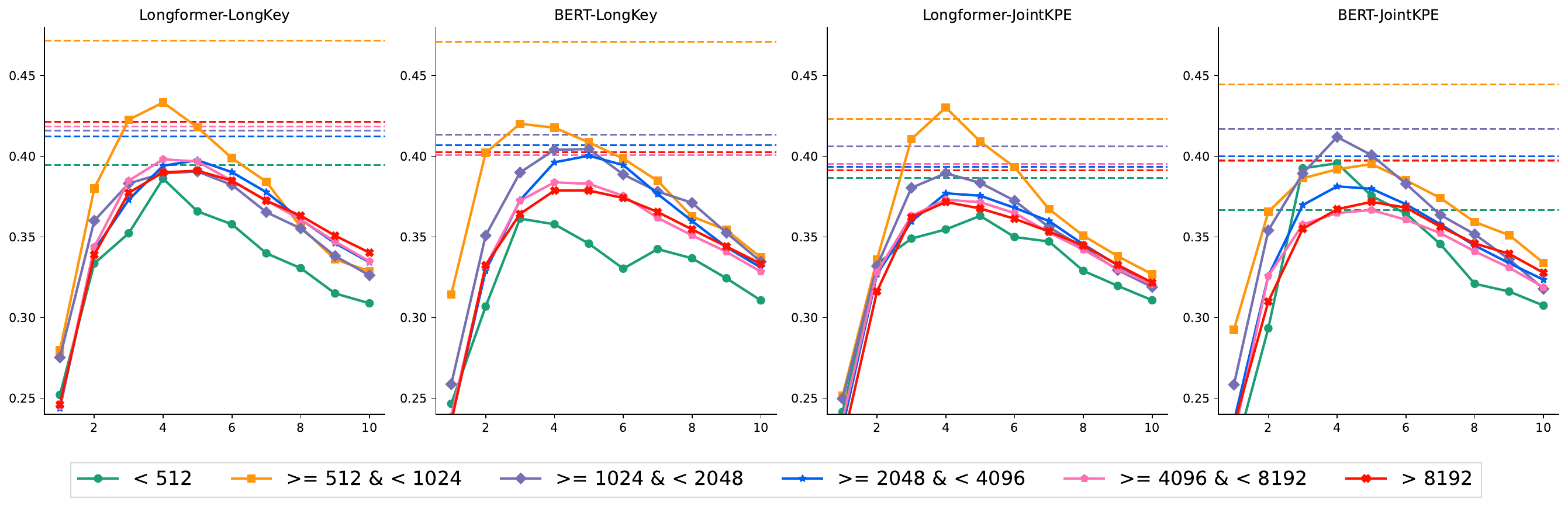}
    \caption{F1 scores based on the document length of the LongKey and JointKPE methods with different encoders applied to the LDKP3K dataset. F1@K for six range of document length (from less than 512 words to more than 8192), where $K=[1,10]$. Dashed lines are the F1@$\mathcal{O}$ for the specific interval.}
    \label{fig:document_length}
\end{figure*}

Figure \ref{fig:document_length} presents the performance of LongKey and JointKPE on the LDKP3K dataset, categorized by document length. Overall, LongKey achieved consistently high scores across different encoder models, while JointKPE's performance was more variable. Notably, LongKey’s Longformer model performed better on longer documents, while the BERT model maintained more balanced results across various lengths. Additionally, LongKey showed particularly strong results for documents between 512 and 1024 tokens, suggesting potential areas for optimization when handling even longer documents.

Overall, LongKey's robustness was evident as it consistently outperformed other models in nearly all benchmarks, showcasing its broad applicability and strength in keyphrase extraction across varied domains. 

\subsection{Component Analysis}

To assess the keyphrase embedding pooler (KEP) contribution, we undertook a component analysis using the LDKP3K validation subset. This analysis involved evaluating the LongKey approach with different aggregation functions, i.e., average, sum and maximum; but also the improvement obtained compared to the JointKPE approach.

We used the configuration outlined in the experimental settings, with each model configuration undergoing 12,500 iterations. Table \ref{tab:component_analysis} shows each configuration's average and standard deviation results that were computed considering five runs per method.

\begin{table}[!htb]
\centering
\caption{Overall results obtained in our component analysis. Scores in \%. The best scores for each K are in \textbf{bold}.}
\label{tab:component_analysis}
\begin{tabular}{@{}lrrr@{}}
\toprule
\multicolumn{4}{c}{\textbf{Component Analysis}} \\ \midrule
\textbf{F1@K} & \textbf{@4} & \textbf{@5} & \textbf{@6} \\ \midrule
JointKPE & 36.03$\pm$0.50 & 36.00$\pm$0.50 & 35.24$\pm$0.48 \\ 
+ avg KEP & 28.60$\pm$0.23 & 29.15$\pm$0.23 & 29.21$\pm$0.34 \\
+ sum KEP & 32.54$\pm$0.36 & 32.76$\pm$0.20 & 32.54$\pm$0.16 \\ \midrule 
LongKey & \textbf{39.04}$\pm$\textbf{0.18} & \textbf{38.94}$\pm$\textbf{0.07} & \textbf{38.13}$\pm$\textbf{0.10} \\ \bottomrule
\end{tabular}
\end{table}

Overall JointKPE F1@5 score was around $36\%$ with a high std dev. of $0.50\%$. Using the KEP proposed in LongKey, but with the average reduction, significantly impaired performance, resulting in an F1@5 score of $29.15\%$, but with a std dev. of $0.23\%$, lower than JointKPE. Using the summation aggregator improved F1@5 a little ($32.76\% \pm 0.20$), but still inferior to JointKPE.

The best F1@5 score was obtained using max pooling, achieving almost $39\%$, with the lowest std. dev. of $0.07\%$.
These findings underscore the KEP's substantial impact on LongKey's success, which is contingent on the appropriate reduction choice.

We suggest that the max aggregator can especially highlight salient features present in different occurrences of a specific keyphrase around the document, thus contributing to a more effective extraction of representative keyphrases. 

\subsection{Performance Evaluation}

We also evaluate the performance of each method in inference. We calculate the performance for each dataset based on the number of processed documents per second using a single RTX 3090. The overall results can be seen in Table \ref{tab:performance_eval}.

\begin{table}[h!t]
\centering
\caption{Performance evaluation of each method tested on each dataset using a single GPU using documents per second. * denotes CPU-only methods.}
\label{tab:performance_eval}
\begin{tabular}{@{}lrrrr@{}}
\toprule
\multicolumn{5}{c}{\textbf{Performance Evaluation (docs/sec)}} \\ \midrule
 & \multicolumn{1}{c}{\textbf{LDKP3K}} & \multicolumn{1}{c}{\textbf{LDKP10K}} & \multicolumn{1}{c}{\textbf{Krapivin}} & \multicolumn{1}{c}{\textbf{SE2010}} \\ \midrule
TF-IDF* & 33.42 & 41.90 & 24.89 & 26.78 \\
TextRank* & 3.70 & 4.26 & 2.57 & 2.39 \\
PatternRank & 1.35 & 1.62 & 1.12 & 1.10 \\ \midrule
SpanKPE & 1.20 & 1.79 & 1.10 & 1.26 \\
TagKPE & 3.96 & 5.06 & 3.02 & 3.03 \\
ChunkKPE & 4.11 & 5.25 & 3.16 & 3.15 \\
RankKPE & 4.16 & 5.26 & 3.22 & 3.19 \\
JointKPE & 4.13 & 5.25 & 3.21 & 3.17 \\
HyperMatch & 4.09 & 5.19 & 3.22 & 3.20 \\ \midrule
BERT-SpanKPE & 0.99 & 1.62 & 0.71 & 0.60 \\
BERT-TagKPE & 5.29 & 6.87 & 3.79 & 3.70 \\
BERT-ChunkKPE & 5.67 & 7.38 & 4.26 & 4.30 \\
BERT-RankKPE & 5.83 & 7.46 & 4.26 & 4.34 \\
BERT-JointKPE & 5.57 & 7.11 & 4.31 & 4.17 \\
BERT-HyperMatch & 5.59 & 7.14 & 4.20 & 4.15 \\ \midrule
\textbf{LongKey} & 4.02 & 5.06 & 3.10 & 3.07 \\
\textbf{BERT-LongKey} & 5.59 & 7.17 & 4.15 & 4.17 \\
\textbf{LongKey8K} & 4.11 & 5.20 & 3.18 & 3.16 \\ \midrule
 & \multicolumn{1}{c}{\textbf{NUS}} & \multicolumn{1}{c}{\textbf{FAO780}} & \multicolumn{1}{c}{\textbf{NLM500}} & \multicolumn{1}{c}{\textbf{TMC}} \\ \midrule
TF-IDF* & 43.35 & 39.79 & 41.11 & 32.50 \\
TextRank* & 4.07 & 2.52 & 3.36 & 2.71 \\
PatternRank & 1.49 & 1.47 & 1.52 & 1.33 \\ \midrule
SpanKPE & 1.08 & 1.20 & 1.56 & 1.29 \\
TagKPE & 4.59 & 4.61 & 4.51 & 3.13 \\
ChunkKPE & 4.82 & 4.82 & 4.71 & 3.20 \\
RankKPE & 4.84 & 4.81 & 4.73 & 3.04 \\
JointKPE & 4.81 & 4.79 & 4.66 & 3.01 \\
HyperMatch & 4.78 & 4.64 & 4.62 & 3.06 \\ \midrule
BERT-SpanKPE & 1.13 & 1.28 & 1.40 & 1.26 \\
BERT-TagKPE & 5.95 & 6.45 & 6.24 & 2.86 \\
BERT-ChunkKPE & 6.70 & 6.78 & 6.38 & 3.98 \\
BERT-RankKPE & 6.59 & 6.90 & 6.54 & 3.67 \\
BERT-JointKPE & 6.57 & 6.89 & 6.46 & 3.49 \\
BERT-HyperMatch & 6.41 & 6.64 & 6.09 & 3.48 \\ \midrule
\textbf{LongKey} & 4.60 & 4.65 & 4.54 & 2.87 \\
\textbf{BERT-LongKey} & 6.42 & 6.59 & 6.23 & 3.44 \\
\textbf{LongKey8K} & 4.70 & 4.73 & 4.66 & 2.96 \\ \bottomrule
\end{tabular}
\end{table}

As we can see, LongKey performed slightly inferior to the supervised methods. This was basically caused by the keyphrase embedding pooler. However, this performance loss is minor compared with how much the overall F1 increased with the proposed module. Robust approaches with as little bottleneck as possible are encouraged. Also, though in some cases BERT-based methods had inferior results, they have a little boost in performance in comparison with Longformer-based.

\subsection{Short Documents}

In Table \ref{tab:short_docs}, we show the results of the evaluated methods in two short-context datasets, KP20k and OpenKP. Three methods were generally competitive: RankKPE, JointKPE, and LongKey. Overall, JointKPE was superior on the KP20k (which was originally developed using it). Since KP20k has a high correlation with LDKP3K, better results are expected in models trained with the latter.

\begin{table*}[h!tb]
\centering
\caption{Results obtained in short document datasets with models trained on LDKP3K and LDKP10K training subsets. Values in \%. Best scores, for each K and dataset, are in \textbf{bold}. Best scores only in a specific section are \underline{underlined}.}
\label{tab:short_docs}
\begin{tabular}{@{}lrrrrrrrrrr@{}}
\toprule
 & \multicolumn{5}{c}{\textbf{KP20k}} & \multicolumn{5}{c}{\textbf{OpenKP}} \\ \midrule
\textbf{F1@K} & \textbf{@3} & \textbf{@4} & \textbf{@5} & \textbf{$\mathcal{O}$} & \textbf{@Best} & \textbf{@3} & \textbf{@4} & \textbf{@5} & \textbf{$\mathcal{O}$} & \textbf{@Best} \\ \midrule
TF-IDF & {\ul 15.43} & {\ul 15.22} & 13.03 & {\ul 15.28} & {\ul 15.45@4} & {\ul 12.48} & {\ul 15.06} & {\ul 13.78} & {\ul 15.17} & {\ul 15.06@3} \\
TextRank & 2.94 & 3.11 & 2.87 & 2.97 & 3.11@5 & 5.39 & 7.54 & 7.56 & 6.86 & 7.70@4 \\
PatternRank & 13.30 & 14.96 & {\ul 14.52} & 12.38 & 15.19@7 & 7.40 & 9.98 & 9.90 & 9.49 & 10.12@4 \\ \midrule
\multicolumn{11}{c}{\textbf{Trained on LDKP3K}} \\ \midrule
SpanKPE & 30.65 & 30.31 & 29.28 & 32.31 & 30.65@3 & 16.87 & 19.35 & 17.84 & 19.88 & 19.41@2 \\
TagKPE & 35.23 & 34.74 & 33.59 & 37.51 & 35.23@3 & 15.93 & 17.42 & 16.06 & 18.21 & 17.52@2 \\
ChunkKPE & 33.66 & 33.11 & 31.98 & 35.88 & 33.66@3 & 16.05 & 18.56 & 17.01 & 18.68 & 18.56@3 \\
RankKPE & 34.77 & 34.45 & 33.35 & 36.76 & 34.77@3 & 16.82 & 20.31 & 18.68 & 20.30 & 20.31@3 \\
JointKPE & \textbf{36.36} & \textbf{35.74} & \textbf{34.45} & \textbf{38.63} & \textbf{36.36@3} & {\ul 17.24} & {\ul 21.25} & {\ul 19.71} & {\ul 20.89} & {\ul 21.26@2} \\ 
HyperMatch & 35.08 & 34.59 & 33.51 & 37.06 & 35.08@3 & 18.58 & 18.09 & 17.41 & 18.20 & 18.58@3 \\ 
\textbf{LongKey} & 35.32 & 35.00 & 33.76 & 37.21 & 35.32@3 & 16.73 & 20.44 & 19.13 & 20.30 & 20.44@3 \\ \midrule
\multicolumn{11}{c}{\textbf{Trained on LDKP10K}} \\ \midrule
SpanKPE & 28.40 & 28.25 & 27.60 & 28.84 & 28.40@3 & \textbf{19.07} & 22.12 & 20.27 & 22.61 & 22.34@2 \\
TagKPE & 28.19 & 28.21 & 27.66 & 29.14 & 28.21@4 & 18.18 & 21.30 & 19.61 & 22.02 & 21.42@2 \\
ChunkKPE & 25.03 & 24.56 & 23.72 & 26.10 & 25.03@3 & 15.57 & 16.51 & 14.80 & 17.38 & 17.07@2 \\
RankKPE & 28.38 & 28.33 & 27.74 & 28.85 & 28.38@3 & 18.79 & \textbf{22.71} & \textbf{21.07} & \textbf{22.86} & \textbf{22.71@3} \\
JointKPE & {\ul 29.20} & 29.06 & 28.37 & 29.84 & 29.20@3 & 17.84 & 22.57 & 21.05 & 22.32 & 22.57@3 \\ 
HyperMatch & 28.02 & 28.35 & 27.79 & 28.14 & 28.35@4 & 20.60 & 20.49 & 19.89 & 20.02 & 20.60@3 \\ 
\textbf{LongKey} & 29.19 & {\ul 29.26} & {\ul 28.65} & {\ul 29.87} & {\ul 29.26@4} & 17.73 & 22.31 & 20.90 & 22.23 & 22.31@3 \\ \bottomrule
\end{tabular}
\end{table*}

For the OpenKP, models trained on the LDKP10K were generally better, especially RankKPE. Here, SpanKPE also had results similar to those of the other three.
Overall, LongKey improvements on long-context datasets (except the TMC dataset, which has a quite different domain) are not seen in short-context documents. These improvements should be related to the proposed keyphrase embedding pooler. Still, LongKey may also be more biased toward long-context documents, which were not generally seen in the training datasets.
Further experiments should be employed, increasing length and content variability in the training stage, to evaluate the capabilities of the keyphrase embedding pooler.

%% file: sections/5conclusion.tex
\section{Conclusion} \label{sec:conclusion}

Automatic keyphrase extraction is crucial for summarizing and navigating the vast content within documents. Yet, prevalent methods fail to analyze long-context texts like books and technical reports comprehensively. To bridge this gap, we introduce LongKey, a novel keyphrase extraction framework specifically designed for the intricacies of extensive documents. LongKey’s robustness stems from its innovative architecture, which is specifically designed for long-form content and rigorously validated on extensive datasets crafted for long-context documents.

To validate its efficacy, we conducted a simple component analysis and further assessments of the LDKP datasets, followed by testing across six diverse and previously unseen long-context datasets and two short-context datasets. The empirical results highlight LongKey's capability in long-context KPE, setting a new benchmark for the field and broadening the horizon for its application across extensive textual domains.

Selecting the appropriate LDKP training dataset was crucial for LongKey's performance on unseen data, highlighting the need for strategic modifications to improve generalization without sacrificing the effectiveness of keyphrase extraction. Slightly inferior results in the short-context datasets also indicate the necessity of improvements for a better generalization.

Furthermore, the restriction on the maximum number of words per keyphrase inherently focuses the method on extracting keyphrases of specific lengths. Further adjustments to accommodate longer keyphrases should be explored, as simply increasing keyphrase length may not improve results without careful evaluation.
Although this is a common pattern in KPE methods, future work must carefully consider the impact of different keyphrase lengths on overall performance.

Additionally, the context size limitation to 8K tokens -- and similarly sized chunks during inference -- may restrict LongKey's ability (through not restricted only to our approach) to fully capture and process extensive document content. However, any plans to expand this limit must carefully balance the increased computational demands with available resources.

In summary, LongKey sets a new benchmark in keyphrase extraction for long documents, combining adaptability with high accuracy across various domains. Its superior embedding strategy contributes to its effectiveness, suggesting significant potential for enhancing document indexing, summarization, and retrieval in diverse real-world contexts.